\documentclass[conference]{IEEEtran}
\IEEEoverridecommandlockouts
\usepackage{cite}
\usepackage{amsmath,amssymb,amsfonts}
\usepackage{algorithmic}
\usepackage{graphicx}
\usepackage{textcomp}
\usepackage{xcolor}
\usepackage{url}
\def\BibTeX{{\rm B\kern-.05em{\sc i\kern-.025em b}\kern-.08em
    T\kern-.1667em\lower.7ex\hbox{E}\kern-.125emX}}
\begin{document}

\title{Word-level Embeddings for Cross-Task Transfer Learning in Speech Processing\\
\thanks{$^1$-These authors contributed equally to this work. We would like to thank Logitech \& Imperial College London for supporting this study.}
}

\author{\IEEEauthorblockN{Pierre Beckmann$^1$}
\IEEEauthorblockA{\textit{Swiss Federal Institute of Technology Lausanne}\\
Lausanne, Switzerland\\
pierre.beckmann@epfl.ch}
\and
\IEEEauthorblockN{Mikolaj Kegler$^1$}
\IEEEauthorblockA{\textit{Imperial College London}\\
London, United Kingdom\\
mikolaj.kegler16@imperial.ac.uk}
\and
\IEEEauthorblockN{Milos Cernak}
\IEEEauthorblockA{\textit{Logitech Europe S.A.}\\
Lausanne, Switzerland\\
milos.cernak@ieee.org}
}

\maketitle

\begin{abstract}
Recent breakthroughs in deep learning often rely on representation learning and knowledge transfer. 
In recent years, unsupervised and self-supervised techniques for learning speech representation were developed to foster automatic speech recognition. Up to date, most of these approaches are task-specific and designed for within-task transfer learning between different datasets or setups of a particular task. In turn, learning task-independent representation of speech and cross-task applications of transfer learning remain less common. Here, we introduce an encoder capturing word-level representations of speech for cross-task transfer learning. We demonstrate the application of the pre-trained encoder in four distinct speech and audio processing tasks: (i) speech enhancement, (ii) language identification, (iii) speech, noise, and music classification, and (iv) speaker identification. In each task, we compare the performance of our cross-task transfer learning approach to task-specific baselines. Our results show that the speech representation captured by the encoder through the pre-training is transferable across distinct speech processing tasks and datasets. Notably, even simple applications of our pre-trained encoder outperformed task-specific methods, or were comparable, depending on the task.
\end{abstract}


\section{Introduction}
\label{sec:intro}

Deep learning frameworks for computer vision and natural language processing often rely on representation learning and knowledge transfer~\cite{tan2018survey}. The goal of transfer learning is to build up domain-specific knowledge on one task and transfer it to another downstream task~\cite{bengio2013representation}. Currently, three main transfer learning approaches can be distinguished: (i) feature extraction, whereby the pre-trained model provides compact representations of domain-specific data~\cite{mesnil2011unsupervised}, (ii) fine-tuning, whereby the knowledge captured by a pre-trained model can be adjusted (i.e. \textit{fine-tuned}) to a particular task or dataset~\cite{ghahremani2017investigation}, and (iii) computing feature losses, whereby representations, obtained through the pre-trained feature extractor, are used to compute losses for training deep learning systems~\cite{dosovitskiy2016generating}.

Following numerous successful applications of deep learning in speech processing, learning representations of speech became the next focus in the field~\cite{chorowski2019unsupervised}.
In particular, learning unsupervised audio representations and evaluating them on downstream classification tasks has recently shown promising results~\cite{google1, google2, google3}. Similarly, one of the recent trends in Automatic Speech Recognition (ASR) is an application of unsupervised~\cite{schneider2019wav2vec,kawakami2020learning} or self-supervised~\cite{baevski2019vq,baevski2020wav2vec} speech representations, as a model pre-training followed by fine-tuning, or as auxiliary speech embedding features.

Recent studies investigating neural encoding of spoken language suggest multi-scale parsing of incoming information into units of the appropriate temporal granularity~\cite{giraud2012cortical}, roughly at a segmental (such as phonetic) and supra-segmental (such as syllabic) timescales. We argue that existing unsupervised and self-supervised speech representations emphasize segmental features (i.e., acoustic models), and when used with a subsequent language model can facilitate ASR. Here, we explore the complementary case, where the acoustic model is represented on the supra-segmental level. Such word-level representation of speech, used without a language model, may be therefore more suitable for a wider range of distinct non-ASR tasks.

In this paper, we introduce a spoken word encoder for learning task-independent representation of speech. In contrast to most current transfer learning approaches, our method was designed to be flexible, versatile and applicable across a range of distinct speech and audio processing tasks. 
Our example encoder presented here adopts VGG-16 architecture~\cite{Simonyan2015VERYRECOGNITION}, similar to VGG-like models  applied previously in audio classification~\cite{hershey2017cnn} and speaker identification~\cite{nagrani2020voxceleb}. Importantly, the proposed methodology does not rely on this particular model, and can be easily employed with other architectures. 

We hypothesize that the spoken word encoder's successive layers capture hierarchically organized generalized representations of speech at the intersection of acoustics and linguistic information. 
Notably, Speech2vec~\cite{chung2018speech2vec} encodes similar spoken word representation, however, it was applied only \textit{within-task} for word similarity experiments, but not other, different downstream tasks.
We thus evaluated our encoder in a \textit{cross-task} and \textit{cross-dataset} configuration using four distinct speech and audio processing problems: (i) speech inpainting~\cite{Kegler2019:Inpainting}, (ii) language identification~\cite{lid}, (iii) speech, noise and music classification~\cite{musan2015} and (iv) speaker identification~\cite{wildermoth,ming,ge}. 

The paper is organized as follows: Section~\ref{sec:methods} introduces the spoken word encoder, its pre-training and its capability for transfer learning. Section~\ref{sec:Experiments} presents four applications of the pre-trained encoder in audio and speech processing tasks, results and comparisons to relevant baselines. Section~\ref{sec:discussion} concludes the paper and outlines avenues for future research.

\section{Spoken word encoder}
\label{sec:methods}

Here we proposed a word encoder based on VGG-16 architecture, thus from now on denoted as \textit{speechVGG}. Importantly, the proposed methods can be paired with other architectures and applied to transfer knowledge between various speech processing tasks beyond those considered here.

\subsection{Encoder architecture}

\begin{figure}[t]
\centering
    \includegraphics[width=\linewidth]{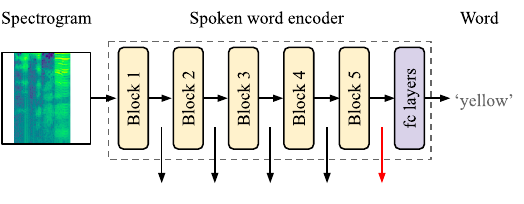}
    \caption{SpeechVGG 
    spoken word encoder and knowledge acquisition via the word classification task. Vertical arrows represent the activation at pooling layers, which reflect the hierarchical representation of speech features captured by the model. The output from the last block provides the most compact representation of speech features (red).
    }
\label{fig:model}
\end{figure}

\begin{figure}[t]
 \centering
  \includegraphics[width=0.9\linewidth]{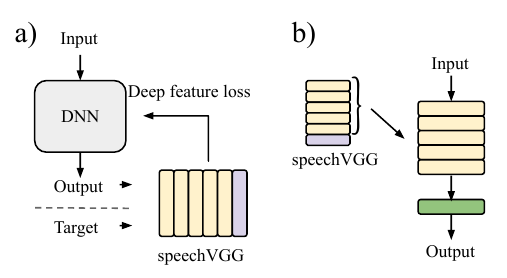}
  \caption{Applications of the spoken word encoder in transfer learning. The pre-trained model can be used to: (a) compute deep feature losses or (b) to transfer knowledge to a new task as a feature extractor (i.e. fixed weights) or as a fully fine-tuneable module.}
 \label{fig:setups}
\end{figure}

Diagram illustrating the architecture of our encoder and the pre-training task is presented in Fig.~\ref{fig:model}. The model adopts the VGG-16 architecture~\cite{Simonyan2015VERYRECOGNITION}. Specifically, the network is built out of five main blocks (Fig.~\ref{fig:model}, yellow), each composed of stacked convolution layers followed by ReLU activation and concluded by a max-pooling layer. The output from the last block is subsequently processed through two fully-connected linear layers followed by a softmax output layer (Fig.~\ref{fig:model}, purple). Note that depending on the task to which the model is deployed, the final fully-connected and output layers of the model may be modified (Fig.~\ref{fig:setups}b).

\subsection{Dataset \& encoder pre-training}
\label{sec:pretrain}

We used LibriSpeech dataset~\cite{Panayotov2015Librispeech:Books} to train the speechVGG. We used all the available training data to build sets of 100 (\textit{train-clean-100}), 460 (+ \textit{train-clean-360}) and 960 hours (+ \textit{train-other-500}) of speech material and used them to train the encoder. We used \textit{test-clean} as a validation set during the model training and \textit{dev-clean} as a separate subset of data to evaluate the performance of the fully trained model. We trained the encoder on the word classification task using different training dictionaries extracted from the LibriSpeech transcriptions. We considered three dictionaries containing 1000, 3000 \& 6000 most frequent, at least 4-letters-long, words from the available data. Together, all considered dictionary and training data sizes made up nine possible training configurations.

For each training setup, we obtained word boundaries (the start and end frames) using forced-alignment from Kaldi LibriSpeech setup~\cite{povey2011kaldi}, and extracted the corresponding segments from the data. We computed log-magnitude spectrograms for each extracted segment by taking absolute values of a complex short-time Fourier transform (STFT, 256 samples window with 128 samples overlap, 128 frequency bins) and then applying natural logarithm. Each frequency channel of the log-magnitude STFT was normalized using mean and standard deviation obtained from the corresponding training dataset. 

Each training sample was augmented using SpecAugment \cite{Park2019SpecAugment:Recognition} to improve the model's generalization capacity. The augmentation was applied by replacing random blocks of time and frequency bins (no more than 50\% in each dimension) in the spectrograms with mean values. To address the varying duration of words, each time-frequency representation of a word was randomly padded with zeros to a size of 128 x 128, corresponding to a 1024-ms-long segment. Such a combination of zero-padding and augmentation facilitated the extraction of speech features in the model. We hypothesize that the zero-padding allows the model to learn to identify parts of the input containing speech, while augmentation makes the learned representations generalized.

Each configuration of the encoder was trained via cross-entropy loss for 30 epochs using ADAM optimizer with a learning rate set to $5\times10^{-5}$. For all of the considered training configurations the model yielded over 92\% accuracy in the word classification task (on held-out data), therefore indicating successful training and knowledge acquisition.

\subsection{Applications in transfer learning}

Our spoken word encoder was designed to extract features from up to 1024-ms-long samples of audio. Features of longer samples can be obtained by averaging the representations from several windows. Due to the hierarchical, modular architecture, each block of the model emphasizes distinct features of the input. The highest-level features, obtained from the last max-pooling in the model, provide the most compact and informative representation of the input (Fig. \ref{fig:model}, red).

The pre-trained encoder can be applied to extract features in a range of speech processing tasks. They can be employed directly as feature extractors in different downstream tasks or used to train deep learning systems via (deep) feature losses~\cite{Kegler2019:Inpainting} (Fig. \ref{fig:setups}a). In the latter case, the extractor pooling layers' activation provides rich representations of both training output and the target. The direct loss computed between these two representations (for example $L_{1}$) can then be used to train the main system.

The pre-trained encoder can be also employed to provide a hot start for learning a brand new task. This can be accomplished by replacing the final layers of the extractor or attaching the other system to the output of the encoder's final block (Fig. \ref{fig:setups}b). Such a system benefits from the knowledge already captured by the pre-trained model and can be furthermore fine-tuned. In the calibration process, the extractor's weights can be either fixed (i.e. \textit{frozen}) or fine-tuned with the rest of the system. In both cases, a set of generalized speech-specific weights facilitates the (re-)training process and the trained system's overall performance.

\section{Transfer learning experiments}
\label{sec:Experiments}

\subsection{Speech inpainting \& benchmarking training setups}

In our previous work~\cite{Kegler2019:Inpainting}, we employed the speechVGG pre-trained using 1000 words from 100 hours of speech recordings to train a deep speech inpainting system for reconstructing missing or distorted parts of the time-frequency representation of speech. Here, we explored how different speechVGG pre-training configurations (section~\ref{sec:pretrain}) influence the speech inpainting performance to determine the optimal setup.

We adopted the exact same framework for deep speech inpainting as introduced in~\cite{Kegler2019:Inpainting}. In particular, we used \textit{train-clean-100} from the LibriSpeech dataset to train the inpainting framework and \textit{dev-clean} as an independent dataset for model evaluation. All the speech material was chunked into 1024-ms-long segments and preprocessed in the same way as for the speechVGG pre-training (see section~\ref{sec:pretrain} for details). Each log-magnitude spectrogram was then distorted using random time \& frequency masks, similar to those applied in SpecAugment \cite{Park2019SpecAugment:Recognition}. The masks removed from 10\% up to 40\% time and frequency bins from the input STFTs. Such samples, along with their mask (i.e. the position of the intrusion was known) were processed through the network to reconstruct the original time-frequency representations of speech. Waveforms were obtained directly from the reconstructed STFT magnitudes using the locally weighted sums algorithm~\cite{LeRoux2010DAFx09}.

The speech inpainting system was trained using nine different configurations of speechVGG specified in  section~\ref{sec:pretrain}. Pre-trained speechVGG was each time used as a feature extractor with fixed weights and applied to compute feature losses for training the inpainting system (Fig.~\ref{fig:setups}a). Specifically, each reconstructed training sample and the corresponding target were processed through the pre-trained speechVGG. The deep feature loss was obtained by computing $L_{1}$ loss between activation of the speechVGG's pooling layers and used to train the speech inpainting model. The inpainting performance of the such trained model was quantified via the short term objective intelligibility (STOI)~\cite{Taal2010ASpeech} and perceptual evaluation of speech quality (PESQ)~\cite{PESQ1998} between the reconstructed and actual speech samples from the held-out dataset (\textit{dev-clean}).

\begin{figure}[h]
\centering
\includegraphics[width=0.65\linewidth]{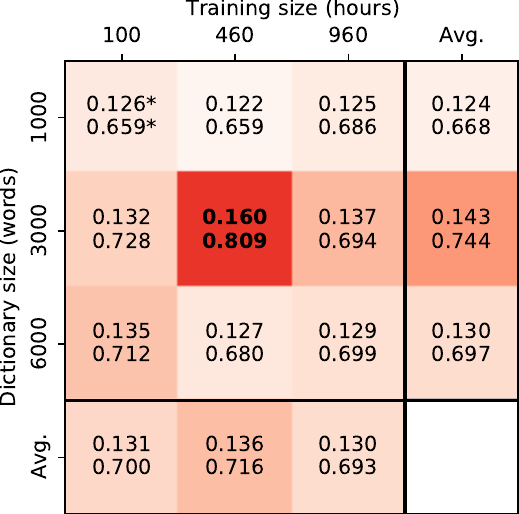}
\caption{Impact of the speechVGG on the training of the speech inpainting system via deep feature loss. Each cell in the array represents improvement of STOI (top) and PESQ (bottom) scores, with respect to the unprocessed case, averaged across the evaluation dataset. *-baseline performance from~\cite{Kegler2019:Inpainting}.}
\label{table:inpainting}
\end{figure}

\textbf{Results}: Improvements of STOI \& PESQ scores through speech inpainting, with respect to the unprocessed, distorted case, are reported in Fig.~\ref{table:inpainting}.
SpeechVGG pre-trained to classify 3000 words extracted from 460 hours of speech recordings was the optimal setup leading to the largest improvements in STOI \& PESQ scores. Notably, this configuration outperformed exsiting baseline employing speechVGG pre-trained to classify 1000 words from 100 hours of speech, as reported in~\cite{Kegler2019:Inpainting}. The lack of improvement for larger sizes of either dictionary or training dataset may be attributed to the fact that over half of the 960 hours of LibriSpeech data belonged to the `other` category, which contains inaccurate annotations of words~\cite{Panayotov2015Librispeech:Books}. We used the best performing configuration of speechVGG (460 hours + 3000 words) for the remaining experiments.

\subsection{Language identification}
\label{sec:lid}

The language identification experiment was performed using the Spoken Language Identification Kaggle dataset~\cite{lid} that contains voice recordings in three languages: English, German, and Spanish. We followed the recommended train/test data split. Twenty randomly chosen 1024-ms-long segments were extracted from each recording and pre-processed to obtain their spectrograms, as specified in section~\ref{sec:pretrain}. Each segment was processed through the pre-trained speechVGG (460 hours + 3000 words), serving as feature extractor with fixed weights, to obtain its representation by flattening the output from the last block (Fig.~\ref{fig:model}, red). A set of features describing a particular recording was each time obtained by averaging representations of its 20 parts. Features obtained from training recordings were used to train a simple logistic regression classifier~\cite{dreiseitl2002logistic} to distinguish the three languages. The trained classifier was evaluated on a separate portion of the data, as stated above.

\textbf{Results}: 
Table~\ref{tab:langid} compares our transfer learning approach with the self-supervised audio representation learning~\cite{google2} and a task-specific convolutional neural network~\cite{lid}. With \textbf{97.6\%} accuracy, our approach outperformed its counterparts using the pre-trained speechVGG as a fixed-weight feature extractor with no additional task-specific fine-tuning of the encoder itself. Although the speechVGG was pre-trained only on English, it was able to accurately distinguish all three languages. This suggests that the representation of speech captured during the encoder pre-training is not language-specific and can generalize to other languages.

\begin{table}[h]
\caption{Language identification task.}
\label{tab:langid}
\centering
\begin{tabular}{l|c}
    \textbf{Method} & \textbf{Accuracy (held-out data)} \\
    \hline
    Tagliasacchi, et al. \cite{google2} & 90.0\% \\
    Task-specific ConvNet~\cite{lid} & 97.0\% \\
    \hline
    \textbf{speechVGG} & \textbf{97.6\%} \\
\end{tabular}
\end{table}

\subsection{Speech, music and noise classification}
\label{sec:musan}

We used the MUSAN dataset~\cite{musan2015} to classify three different categories of audio recordings: speech (recordings from the US government and librivox.org), music, and noise. We discarded all audio samples shorter than 1024 ms from the dataset. All such short samples were recordings of noise, and including them could lead to biased predictions based solely on the sample duration rather than its acoustic content. From the remaining data we set aside randomly selected 10\% as a held-out evaluation set. Analogously to the previous task (section~\ref{sec:lid}), twenty randomly chosen 1024-ms-long segments were obtained from each recording in the dataset and pre-processed as specified in section~\ref{sec:pretrain}. The pre-trained speechVGG (460 hours + 3000 words), with fixed weights and no additional fine tuning, was used to obtain features from each segment. Same as before, a set of features for each recording was obtained by averaging representation of its segments. The features from the training portion of the data were used to train a simple logistic regression classifier~\cite{dreiseitl2002logistic} to distinguish speech, music, and noise. The classifier was evaluated using samples from the held-out portion of the data.

\begin{figure}[h]
\centering
  \includegraphics[width=\linewidth]{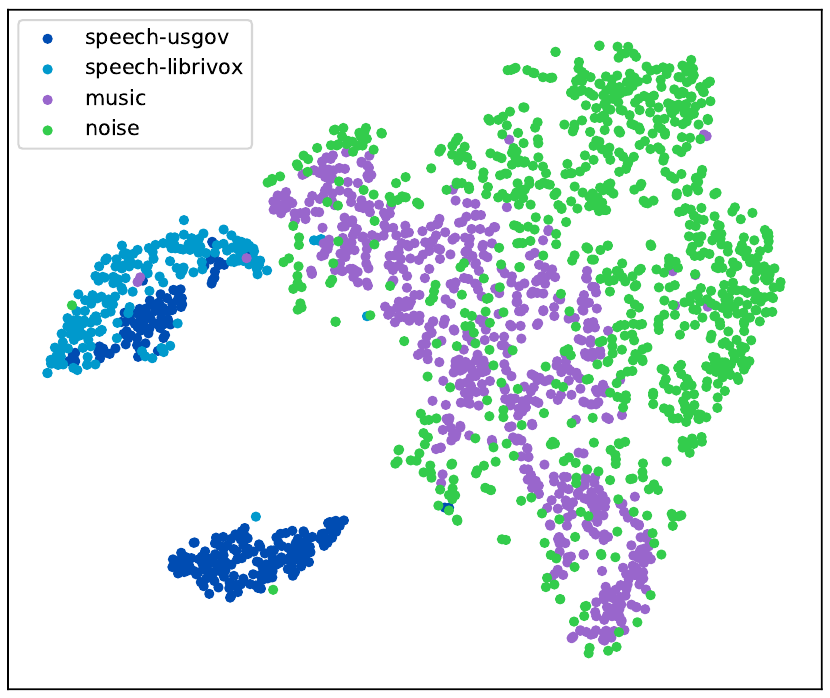}
  \caption{t-SNE visualization of high-dimensional embeddings of speech, music and noise recordings from MUSAN, obtained via the pre-trained speechVGG.}
  \label{fig:scatter}
\end{figure}

\textbf{Results}: High-dimensional embeddings of the MUSAN dataset were visualized via t-SNE~\cite{maaten2008visualizing} (Fig.~\ref{fig:scatter}). Clusters representing speech recordings were clearly distinguishable from music and noise. Interestingly, embeddings of speech recordings were divided into two distinct clusters; one made up almost exclusively of US government recordings (Fig.~\ref{fig:scatter}, dark blue). Our approach yielded \textbf{96.5\%} classification accuracy. This suggests that speechVGG, pre-trained on the LibriSpeech data, not only successfully transferred the generalized speech representation to this task, but also allowed to reliably distinguish samples of music and noise. Tagliasacchi, et al.~\cite{google2} reported 99.0\% accuracy on this task using representations from 0.975-seconds-long segments of recordings. Importantly, their approach was tailored specifically for this task, while our encoder was designed for versatile \textit{cross-task} applications. Moreover, our approach used only twenty 1024-ms-long segments from each clip, i.e. up to 20.5 seconds of audio, instead of all available, as Tagliasacchi, et al.~\cite{google2} did.

\subsection{Speaker identification}
\label{subsec:speakers}

In the speaker identification task, we used speech recordings from the TIMIT dataset, including 630 speakers \cite{timit}. We randomly selected one recording per speaker to form a set-aside evaluations set, while the rest of the data was used for training. All the data was chunked into 1024-ms-long segments, and the log-magnitude spectrogram of each chunk was obtained as specified in section~\ref{sec:pretrain}. The previously introduced approach, where the pre-trained speechVGG (460 hours + 3000 words) was used as a feature extractor with fixed weights, did not succeed in this task and led to poor performance. We thus replaced the output layer of the speechVGG to accommodate a different number of classes in the new task (630 speakers) and fine-tuned the model (Fig.~\ref{fig:setups}b). For fine tuning, we used the same training routines as in the speechVGG pre-training, but fed the model with the task-specific data. In particular, the model was trained to classify speakers based on a single 1024-ms-long window. During evaluation on a set-aside portion of the data, the speaker identity was determined by averaging model predictions from a window sliding over the entire recording with a 50\% overlap (512 ms).

\textbf{Results}: Results of speaker identification using the fine-tuned speechVGG, alongside the baseline approaches, are presented in Table~\ref{tab:speaker}. The fine-tuned speechVGG achieved \textbf{99.7\%} accuracy on the set-aside portion of the data and therefore outperformed existing methods evaluated on the entire TIMIT corpus~\cite{ming,wildermoth}. Ge et al. (2017)~\cite{ge} reported 100\% accuracy in this task employing a 1-second-long window using a subset of 100 male speakers. In the same setup our fine-tuned model also solved the task yielding \textbf{100\%} accuracy (Table~\ref{tab:speaker} - 100 male speakers).

\begin{table}[h]
\caption{Speaker identification task.}
\label{tab:speaker}
\centering
\begin{tabular}{l|ccll}
\multicolumn{1}{c|}{\textbf{}} & \multicolumn{2}{c}{\textbf{Accuracy (held-out data)}} &  &  \\
\multicolumn{1}{c|}{\textbf{Method}} & \multicolumn{1}{c|}{\textbf{All speakers}} & \textbf{100 male speakers} &  &  \\ \cline{1-3}
Ming, et al.~\cite{ming} & \multicolumn{1}{c|}{608/630 (96.5\%)} & - &  &  \\
Wildermoth, et al.~\cite{wildermoth} & \multicolumn{1}{c|}{623/630 (99.0\%)} & - &  &  \\
Ge, et al.~\cite{ge} & \multicolumn{1}{c|}{-} & \textbf{100/100 (100\%)} &  &  \\ \cline{1-3}
\textbf{speechVGG} (fine-tuned) & \multicolumn{1}{c|}{\textbf{628/630 (99.7\%)}} & \textbf{100/100 (100\%)} &  & 
\end{tabular}
\end{table}

\section{Discussion}
\label{sec:discussion}

Here, we proposed an approach for learning word-level embeddings, suitable for flexible knowledge transfer across different speech and audio processing tasks. In contrast to most existing \textit{task-specific} transfer learning approaches our method is focused on versatility and \textit{cross-task} compatibility. We evaluated the proposed spoken word encoder as a `general-purpose` speech feature extractor and explored its performance in a range of distinct speech and audio processing tasks. 

The generalized representation of speech captured during the encoder pre-trained on the LibriSpeech dataset~\cite{Panayotov2015Librispeech:Books} was transferable over four distinct tasks, employing different datasets, not limited to speech~\cite{lid,musan2015,timit}. Interestingly, relatively simple applications of our pre-trained spoken word encoder were capable of achieving results comparable to the recent task-specific approaches with little to no additional fine-tuning (section~\ref{sec:Experiments}). Implementation of the speechVGG, pre-trained models and example applications are available at\footnote{\url{https://github.com/bepierre/SpeechVGG}}.

We would like to re-iterate that the proposed pre-training and transfer learning methodology is not restricted to the example encoder architecture introduced in Section~\ref{sec:methods}. Depending on the tasks of interest, the feature extractor can be of considerably higher or lower complexity than the example presented here. In particular, systematic exploration of different encoder configurations and fusion of our approach with existing (self-)supervised and unsupervised training setups may further improve efficacy of the proposed framework. 

\bibliographystyle{ieeetr}
\bibliography{references}

\end{document}